\documentclass[letterpaper]{article} 
\usepackage{aaai2026}  
\usepackage{times}  
\usepackage{helvet}  
\usepackage{courier}  
\usepackage[hyphens]{url}  
\usepackage{graphicx} 
\usepackage{natbib}  
\usepackage{caption} 
\usepackage{algorithm}
\usepackage{algorithmic}

\usepackage{amsmath}
\usepackage{amsthm}
\usepackage{booktabs}
\usepackage{multirow}
\usepackage{amssymb}
\usepackage{newfloat}
\usepackage{listings}
\DeclareCaptionStyle{ruled}{labelfont=normalfont,labelsep=colon,strut=off} 
\floatstyle{ruled}
\newfloat{listing}{tb}{lst}{}
\floatname{listing}{Listing}
\title{Differentiable Sparse Identification of Lagrangian Dynamics}

\author {
    Zitong Zhang,
    Hao Sun\thanks{Corresponding author.}
}
\affiliations {
    Gaoling School of Artificial Intelligence, Renmin University of China, Beijing, China\\
    \{zhangzitong, haosun\}@ruc.edu.cn
}
\begin{document}

\maketitle

\begin{abstract}
Data-driven discovery of governing equations from data remains a fundamental challenge in nonlinear dynamics. Although sparse regression techniques have advanced system identification, they struggle with rational functions and noise sensitivity in complex mechanical systems. The Lagrangian formalism offers a promising alternative, as it typically avoids rational expressions and provides a more concise representation of system dynamics. However, existing Lagrangian identification methods are significantly affected by measurement noise and limited data availability. This paper presents a novel differentiable sparse identification framework that addresses these limitations through three key contributions: (1) the first integration of cubic B-Spline approximation into Lagrangian system identification, enabling accurate representation of complex nonlinearities, (2) a robust equation discovery mechanism that effectively utilizes measurements while incorporating known physical constraints, (3) a recursive derivative computation scheme based on B-spline basis functions, effectively constraining higher-order derivatives and reducing noise sensitivity on second-order dynamical systems. The proposed method demonstrates superior performance and enables more accurate and reliable extraction of physical laws from noisy data, particularly in complex mechanical systems compared to baseline methods.
\end{abstract}


\section{Introduction}
\label{sec:intro}
Nonlinear dynamics plays a pivotal role across scientific and engineering domains, such as chaotic weather systems, celestial mechanics, neural oscillations, etc. Deriving governing equations directly from data has emerged as a transformative approach for understanding and predicting such complex systems.
The autonomous discovery of physical laws from data has been a long-standing goal in scientific research. 
Deep learning provides powerful tools for modeling dynamical systems and addressing complex challenges~\cite{battaglia2016interaction,lenz2015deepmpc,sahoo2018learning}. These models excel at approximating some fundamental physical principles,including Hamiltonians~\cite{greydanus2019hamiltonian} and Lagrangians~\cite{cranmer2020lagrangian}, as well as sophisticated mathematical frameworks such as Koopman eigenfunctions~\cite{lusch2018deep}. 
These eigenfunctions transform nonlinear dynamics into a linear framework within an infinite-dimensional Hilbert space~\cite{koopman1932dynamical,kaiser2021data}, offering a powerful tool for the analysis and control of systems that are otherwise computationally intractable. 
Recent work has even combined GNNs with Koopman analysis to model molecular dynamics in an unsupervised manner~\cite{xie2019graph}.
While deep learning models often lack interpretability, and function as black boxes, obscuring the underlying relationships between variables and providing limited insight into the mechanisms governing these relationships, interpretable models provide a compelling alternative. 
By offering transparent and causally interpretable insights into system behavior, these models enable reliable predictions and a profound understanding of complex dynamics.
These approaches bridge a critical gap by providing methods that combine accuracy with interpretability.

Symbolic learning methods~\cite{bongard2007automated,quade2016prediction} seek to extract interpretable mathematical models from data. A significant advancement in this field was achieved by ~\cite{schmidt2009distilling}, who introduced symbolic regression to distill physical laws from experimental data while balancing accuracy and interpretability. This approach provides a principled framework for discovering concise representations of natural phenomena. However, it scales poorly to high-dimensional systems due to the exponential growth of the search space, limiting its practical applicability.
An alternative direction explores symbolic neural networks~\cite{martius2016extrapolation,sahoo2018learning,long2019pde,petersen2020deep}, which employ mathematical operators as activations to create interpretable models~\cite{mundhenk2021symbolic,sun2023symbolic}. While weight pruning enables parsimony, their reliance on user-defined thresholds compromises robustness. Furthermore, numerical differentiation of system responses exacerbates sensitivity to noise and sparse data, limiting practical applicability.

The Sparse Identification of Nonlinear Dynamics (SINDy)~\cite{brunton2016discovering,rudy2017data} method has emerged as a powerful framework for discovering governing equations from data. By constructing a library of candidate nonlinear terms and applying sparse regression techniques. Using techniques like sequential threshold ridge regression (STRidge), SINDy iteratively refines sparse solutions through adaptive thresholding, enabling the discovery of interpretable and parsimonious models. While SINDy has demonstrated broad applicability, it struggles with dynamics involving rational functions due to the exponential growth of the candidate function library, complicating sparse regression.
SINDy-PI~\cite{kaheman2020sindy} improves noise robustness through convex optimization but is limited to low noise levels and risks instability if incorrect denominator terms are identified. Similarly, RK4-SINDy~\cite{goyal2022discovery}, which combines Runge-Kutta integration with sparse identification, has shown promise for simple rational systems but may fail in complex scenarios due to potential errors in denominator selection, leading to unstable predictions. These limitations call for more robust methods to handle rational functions.
Physics-informed learning helps recover system states and estimate derivatives from sparse, noisy data for equation discovery. Methods like neural networks~\cite{sun2022bayesian} and cubic splines~\cite{sun2021physics, zhang2024vision} are commonly used.

Lagrangian provides a compact representation of system dynamics, encapsulating all predictive information within a single scalar function, unlike the more complex structure of differential equations.
Lagrangian-SINDy~\cite{chu2020discovering} formulates the total energy as a Koopman eigenfunction with zero eigenvalue and applies sparse regression to align the time derivative of this energy with the system's net power input, enabling accurate recovery of the Lagrangian structure.
xL-SINDy~\cite{purnomo2023sparse}, an extension of Lagrangian-SINDy, integrates the SINDy framework with the proximal gradient method to derive sparse Lagrangian representations.
However, current Lagrangian-based identification methods remain limited by their susceptibility to noise or their dependence on substantial data availability due to the chain rule differentiation required for library functions.
\begin{figure*}[t]
  \centering
  \includegraphics[width=0.95\linewidth]{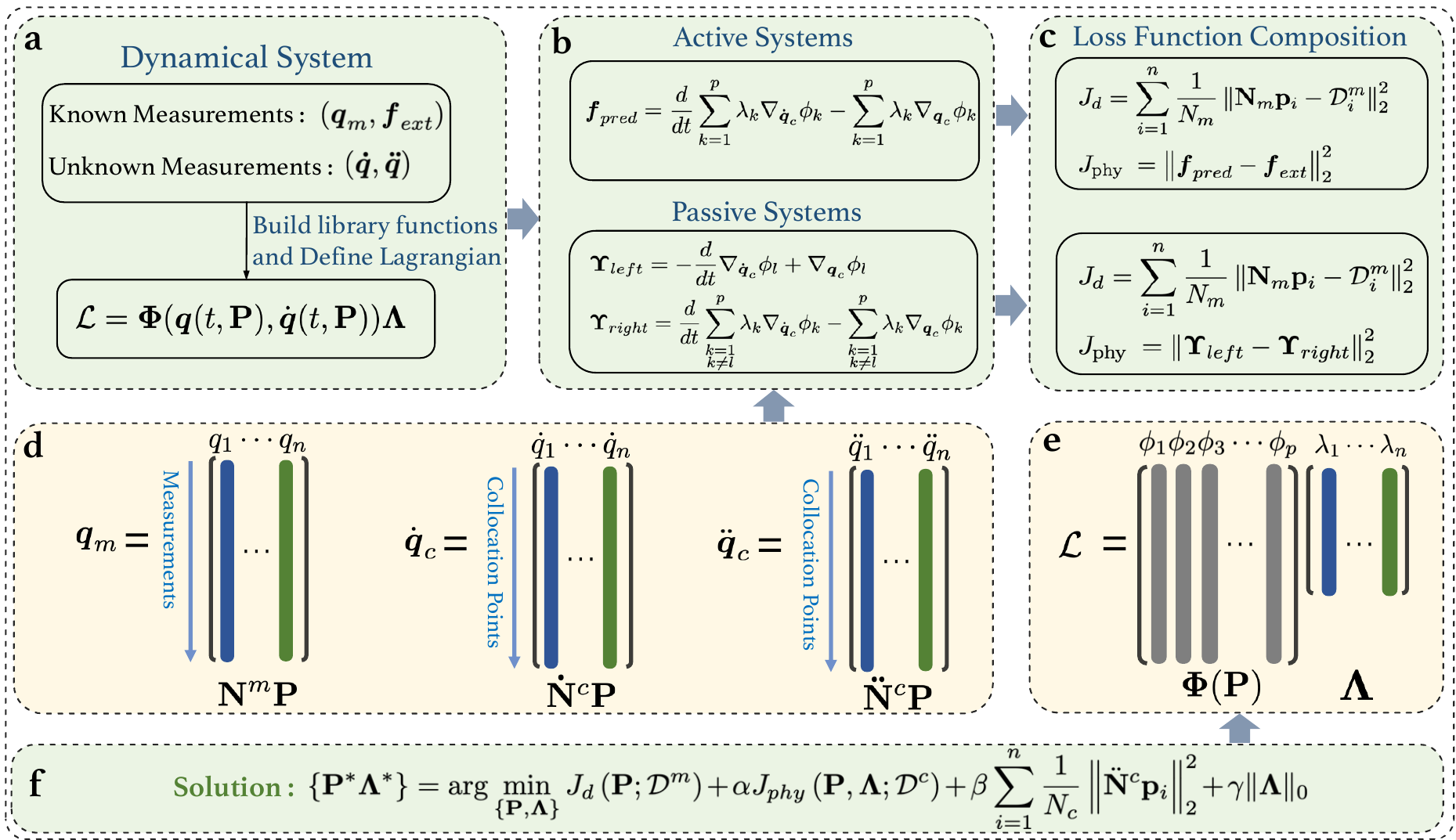}
  \caption{
  Schematic architecture of differentiable sparse identification of lagrangian dynamics. 
    (a) Measurement of the dynamical system and definition of the Lagrangian equations; 
    (b) Description of equation formulations across different systems; 
    (c) Loss functions under various system configurations; 
    (d) Representation of values and their derivatives (first and second order) using cubic B-splines for each degree of freedom; 
    (e) Schematic illustration of the sparse representation of the Euler-Lagrange equations; 
    (f) Network is equivalent to solving optimization problem.
  }
  \label{fig: architecture}
\end{figure*}
\paragraph{Contributions:} The key contributions of this work are as follows:
(1) We bridge this gap by introducing a novel framework that combines Cubic B-Spline approximation with sparse regression for Lagrangian system identification, providing a flexible framework for accurately modeling complex nonlinearities.
(2) We address critical limitations of existing Lagrangian-based methods by significantly reducing their sensitivity to noise and reliance on large data volumes. This is achieved through a novel formulation that minimizes the adverse effects of chain rule differentiation on library functions.
(3) We develop a robust equation discovery framework that efficiently leverages limited measurements while incorporating physical constraints, enabling precise reconstruction of governing equations via sparse regression. 
By leveraging the recursive properties of B-spline basis functions, we enable robust derivative computation and noise mitigation, particularly for second-order dynamical systems. This advancement opens new possibilities for extracting physical laws from measurements, with potential applications in robotics, biomechanics, and beyond.

\section{Methodology}
We introduce a framework for differentiable sparse identification of lagrangian dynamics. Figure~\ref{fig: architecture} depicts the overall architecture of our method. 

\subsection{Cubic B-Spline Approximation}


B-splines are piecewise polynomial basis functions with inherent differentiability. 
Cubic B-splines provide a differentiable surrogate framework, offering a flexible approach for curve fitting.
Since the first derivative of a B-spline curve is another B-spline curve, higher-order derivatives can be computed recursively by applying this property. Consequently, the first and second derivatives of each basis function can be efficiently obtained(see Appendix~\ref{app:sec_derivative} for derivations). 

To ensure the curve possesses continuous and smooth tangent directions at the starting and ending nodes, meeting the first derivative interpolation requirement, we use clamped B-spline curves for fitting. 
This approach enforces the curve to pass through the endpoints by repeating the boundary knots \(p+1\) times, where \(p\) is the degree of the B-spline, ensuring both endpoint interpolation and smoothness across the curve.
In the context of B-spline curves, the term \( P \) denotes the control points, which are a set of points used to define the shape of the curve. The first and last control points are set to the start and end points of the data, respectively:
\(
P_0 = D_0, P_s = D_m,
\)
where $m$ and $s$ denote the number of data points and control points, and \( D_0 \) and \( D_m \) are the first and last data points. For a B-spline of degree \( p \), the first and last knots are repeated. For example, in the case of a cubic B-spline (\( p = 3 \)), the knot vector \( U \) is constructed as:
\(
U = [u_0, u_0, u_0, u_0, u_1, u_2, \dots, u_{r-1}, u_r, u_r, u_r, u_r].
\)
This ensures sufficient support for the boundary basis functions. Mathematically, a B-spline curve \( C(u) \) of degree \( p \) is expressed as a weighted sum of control points \( P_i \) and basis functions \( N_{i,p}(u) \):
\begin{equation}
    C(u) = \sum_{i=0}^{s} N_{i,p}(u) \cdot P_i,
\end{equation}
where  \( P_i \) represents the \( i \)-th control point, \( N_{i,p}(u) \) is the \( i \)-th B-spline basis function of degree \( p \), \( u \) is the parameter within the knot vector \( U \).
The control points \( P_i \) play a crucial role in determining the geometry of the curve. By adjusting their positions, the shape of the B-spline curve can be precisely controlled. Additionally, the use of clamped boundary conditions ensures that the curve interpolates specific control points, such as the start point \( P_0 \) and the end point \( P_s \).

\subsection{Network Architecture}
\paragraph{Formulation of Lagrangian Acquisition for Dynamical Systems:}

For a multidegree-of-freedom nonrelativistic system, the Lagrangian $\mathcal{L}$ is given by:
\begin{equation}
\label{eq: lagrangian_Eq_companent}
    \mathcal{L} = \sum_i (T_i - V_i),
\end{equation}
where $T_i$ and $V_i$ denote the kinetic and potential energies of the $i$-th component, respectively. Given that the total Lagrangian is the sum of the Lagrangians of its individual components, any nonlinear terms appearing in the Lagrangian of a component will inherently be reflected in the total Lagrangian of the system.

For a system with \(n\) degrees of freedom, we formulate the Lagrangian as a linear combination of nonlinear candidate functions. Let $\boldsymbol{q}=(q_1, q_2, \dots, q_n)$ denote the generalized coordinates of the system. 
\(q_i(t)\) is a function describing the time evolution of the \(i\)-th generalized coordinate. 
The Euler-Lagrange equations govern the dynamics of the Lagrangian system, as described by:

\begin{equation}
\label{eq: lagrangian_Eq}
    \mathcal{L}=\sum_{k=1}^p \lambda_k \phi_k(\boldsymbol{q}, \dot{\boldsymbol{q}}) = \boldsymbol{\Phi}(\boldsymbol{q}, \dot{\boldsymbol{q}})\boldsymbol{\Lambda}.
\end{equation}
Here, $\boldsymbol{\Phi}(\boldsymbol{q}, \dot{\boldsymbol{q}})$ is the library of nonlinear candidate functions, with each column being a nonlinear candidate function $\phi_k(\boldsymbol{q}, \dot{\boldsymbol{q}})$. The vector $\boldsymbol{\Lambda} = (\lambda_1, \lambda_2, \dots, \lambda_n)^\top$ contains the coefficients weighting each function in the Lagrangian $\mathcal{L}$.

We investigate two distinct scenarios: (1) active systems with external input \( \boldsymbol{f}_{ext} \), and (2) passive systems where some prior knowledge of the Lagrangian is available. 
Schematic architecture of our approach and problem formulation for each scenario is summarized in Figure~\ref{fig: architecture}.
 
\paragraph{Active dynamical systems} Active systems are characterized by non-conservative forces (e.g., control inputs or driving forces) through external input \(\boldsymbol{f}_{ext} = (f_1, f_2, \dots, f_n)^\top\). The equation for active systems is:
\begin{equation}
\label{eq: lagrangian_Eq_active}
    \frac{d}{dt} \nabla_{\dot{\boldsymbol{q}}} \mathcal{L}  - \nabla_{\boldsymbol{q}} \mathcal{L} = \boldsymbol{f}_{ext},
\end{equation}
where \(\nabla_{{\dot{\boldsymbol{q}}}} \equiv \frac{\partial }{\partial \dot{\boldsymbol{q}}}\) and \(\nabla_{{{\boldsymbol{q}}}} \equiv \frac{\partial }{\partial {\boldsymbol{q}}}\).
Incorporating Eq.~(\ref{eq: lagrangian_Eq}) into Eq.~(\ref{eq: lagrangian_Eq_active}) for active systems gives:
\begin{equation}
\label{eq: lagrangian_Eq_active_pred}
\boldsymbol{f}_{pred}=\frac{d}{d t} \sum_{k=1}^p \lambda_{k} \nabla_{\dot{\boldsymbol{q}}} \phi_k-\sum_{k=1}^p \lambda_{k} \nabla_{\boldsymbol{q}} \phi_k
\end{equation}

where \(\boldsymbol{f}_{pred}\) denote the predicted value of the external input $\boldsymbol{f}_{ext}$ based on a coefficient vector $(\lambda_1, \lambda_2, \dots, \lambda_n)$. The time derivative $\frac{d}{dt}$ can be further expanded using the chain rule, yielding terms that involve $\dot{\boldsymbol{q}}$ and $\ddot{\boldsymbol{q}}$ as follows:
\begin{equation}
\label{eq: lagrangian_Eq_active_pred_expand}
\begin{aligned}
    \boldsymbol{f}_{pred}
    & =\sum_{k=1}^p \lambda_k\left(\nabla_{\dot{\boldsymbol{q}}}^{\top} \nabla_{\dot{\boldsymbol{q}}} \phi_k \ddot{\boldsymbol{q}}+\nabla_{\boldsymbol{q}}^{\top} \nabla_{\dot{\boldsymbol{q}}} \phi_k \dot{\boldsymbol{q}}-\nabla_{\boldsymbol{q}} \phi_k\right)
\end{aligned}
\end{equation}

\paragraph{Passive dynamical systems} Passive systems rely solely on internal energy (e.g., potential and kinetic energy) without external input. They are conservative systems where energy is conserved, and the generalized force $f_i$ is zero:
\begin{equation}
\label{eq: lagrangian_Eq_passive}
    \frac{d}{dt} \left( \frac{\partial \mathcal{L}}{\partial \dot{\boldsymbol{q}}} \right) - \frac{\partial \mathcal{L}}{\partial \boldsymbol{q}} = 0.
\end{equation}

\noindent
For passive systems, certain terms in the Lagrangian equation can still be determined based on prior knowledge of the system, such as the mass of a component, gravitational constant, elastic coefficient, length, and mass of objects. Furthermore, as indicated by Eq.~(\ref{eq: lagrangian_Eq_companent}), the Lagrangian of the system is constructed as the sum of the kinetic energy minus the potential energy for all components in the system. Since the Lagrangian is not unique, we normalize the coefficients of the known terms to unity for simplicity. Equation~(\ref{eq: lagrangian_Eq}) can be expressed as follows:
\begin{equation}
\label{eq: lagrangian_Eq_passive_prior}
    \mathcal{L}=\phi_l(\boldsymbol{q}, \dot{\boldsymbol{q}})+\sum_{\substack{k=1 \\ k \neq l}}^p \lambda_k^{} \phi_k(\boldsymbol{q}, \dot{\boldsymbol{q}}).
\end{equation}
\noindent
Therefore, we can extract prior knowledge of a constituent of the system and focus on identifying the coefficients of the remaining terms.

\paragraph{Network Architecture Description:}
The network architecture overview, as illustrated in Figure~\ref{fig: architecture}. Initially, we define $n$ sets of control points for cubic B-splines, represented as $\mathbf{P} = \{\mathbf{p}_1, \mathbf{p}_2, \dots, \mathbf{p}_n\} \in \mathbb{R}^{s \times n}$. These control points are then combined with the spline basis functions $\mathbf{N}(t)$ to interpolate the system's $n$-dimensional state of system: 
\(\mathbf{q}(t, \mathbf{P}) = \mathbf{N}(t) \mathbf{P}.\)
Similarly, the first and second derivatives of the generalized coordinates can be expressed as:
\(
\dot{\mathbf{q}}(t, \mathbf{P}) = \dot{\mathbf{N}}(t) \mathbf{P},
\) 
\(
\ddot{\mathbf{q}}(t, \mathbf{P}) = \ddot{\mathbf{N}}(t) \mathbf{P}
\).
The Lagrangian $\mathcal{L}$ in Eq.~\ref{eq: lagrangian_Eq} can be rewritten in the following form:
\begin{equation}
\label{eq: lagrangian_Eq_aboutP}
    \mathcal{L}= \boldsymbol{\Phi}(\boldsymbol{q}(t, \mathbf{P}), \dot{\boldsymbol{q}}(t, \mathbf{P}))\boldsymbol{\Lambda}.
\end{equation}
The discovery problem can be formally defined as follows: given the dataset $\mathcal{D}^m = \{ \boldsymbol{q}_m^i \}_{i=1,\dots,n} \in \mathbb{R}^{N_m \times n}$, the goal is to determine the optimal parameters $\mathbf{P}$ and $\boldsymbol{\Lambda}$ such that Eq.~(\ref{eq: lagrangian_Eq_aboutP}) is satisfied for $\forall t$. Here, $N_m$ represents the number of measurement points. 
$\mathcal{D}^c = \{t_0, t_1, \dots, t_{N_c-1}\}$ represents a set of $N_c$ randomly sampled collocation points, where $N_c \gg N_m$. These points are used to improve the satisfaction of physical constraints.
The matrix $\mathbf{N}_m \in \mathbb{R}^{N_m \times s}$ represents the spline basis matrix evaluated at the measured time instances, while $\dot{\mathbf{N}}_c \in \mathbb{R}^{N_c \times s}$ denotes the derivative of the spline basis matrix evaluated at the collocation instances.

The loss function of the proposed network consists of two components: the $\textit{data loss}$ (${J}_d$) and the $\textit{physics loss}$ (${J}_{phy}$). As illustrated in Figure~\ref{fig: architecture}$\mathbf{c}$, while the data loss is formulated identically for both cases, the physics loss differs between active and passive systems. For active systems, the physics loss minimizes the difference between the predicted external force $\boldsymbol{f}_{{pred}}$ and the known external force $\boldsymbol{f}_{{ext}}$. For passive systems, it minimizes the discrepancy between the terms involving prior knowledge $\boldsymbol{\Upsilon}_{{left}}$ and the terms to be determined $\boldsymbol{\Upsilon}_{{right}}$.
Mathematically, training the network is equivalent to solving the optimization problem (see Figure~\ref{fig: architecture}$\mathbf{f}$),
where $\alpha$, $\beta$, and $\gamma$ are weighting coefficients. 
To promote optimization, constrain higher-order derivatives, and mitigate noise amplification caused by chain rule differentiation---which exacerbates noise sensitivity in second-order dynamical systems---we introduce the regularization term $J_{reg} = \beta \sum_{i=1}^n \frac{1}{N_c} \|\ddot{\mathbf{N}}^c \mathbf{p}_i\|_2^2$, ensuring better alignment with physical principles and numerically stable.

Inspired by the STLS (Sequential Thresholded Least-Squares) approach in SINDy, we adopt an iterative procedure to identify sparse dynamics. The process begins by fitting the coefficients using ordinary least squares. A thresholding step is then applied to eliminate terms with coefficients below a predefined threshold, effectively enforcing sparsity. The algorithm iterates by refitting the remaining terms and reapplying the threshold until convergence is achieved. This approach ensures that only the most significant terms are retained in the final model, resulting in a parsimonious representation of the system's dynamics. By significantly reducing the number of candidate functions considered during learning, this step accelerates convergence compared to methods without hard thresholding. Finally, we verify whether the cost function has reached the specified tolerance.
By substituting the optimal solutions $\mathbf{P}^*$ and $\boldsymbol{\Lambda}^*$ into Eq.~(\ref{eq: lagrangian_Eq_aboutP}), we obtain the Lagrangian of the dynamical system. 
This overall training scheme, including initialization, optimization loops, and convergence criteria, is systematically outlined in Algorithm~\ref{alg:lagrangian_discovery}.
\begin{algorithm}[tb]
\caption{Differentiable Sparse Identification of Lagrangian Dynamics}
\label{alg:lagrangian_discovery}
\textbf{Input}: $\mathcal{D}^m$,
$\mathcal{D}^c$, \( \boldsymbol{f}_{ext} \) or known terms\\
\textbf{Parameter}: Threshold $\epsilon$, weights $\alpha,\beta,\gamma$\\
\textbf{Output}: Discovered \(\mathcal{L}^*\)
\begin{algorithmic}[1]
\STATE Construct nonlinear candidate function library$\boldsymbol{\Phi}$. Initialize B-spline control points $\mathbf{P}$ and library coefficients $\boldsymbol{\Lambda}$
\STATE Construct basis matrices $\mathbf{N}_m$, $\dot{\mathbf{N}}_c$, $\ddot{\mathbf{N}}_c$
\REPEAT
\STATE Compute: $\mathbf{q} = \mathbf{N}_m\mathbf{P}$, $\dot{\mathbf{q}} = \dot{\mathbf{N}}_c\mathbf{P}$, $\ddot{\mathbf{q}} = \ddot{\mathbf{N}}_c\mathbf{P}$
\STATE Evaluate losses:
\STATE \quad $J = J_d + J_{phy} + J_{reg}$
\STATE Update $\mathbf{P}, \boldsymbol{\Lambda}$ via gradient descent on $J$
\STATE Apply STLS thresholding: $\Lambda_{ij} \gets 0$ if $|\Lambda_{ij}| < \epsilon$
\UNTIL{convergence}
\STATE \textbf{return} $\mathcal{L}^* = \boldsymbol{\Phi}(\mathbf{q},\dot{\mathbf{q}})\boldsymbol{\Lambda}^*$
\end{algorithmic}
\end{algorithm}

\section{Experiments} 
\begin{figure} 
  \centering
  \includegraphics[width=0.98\linewidth]{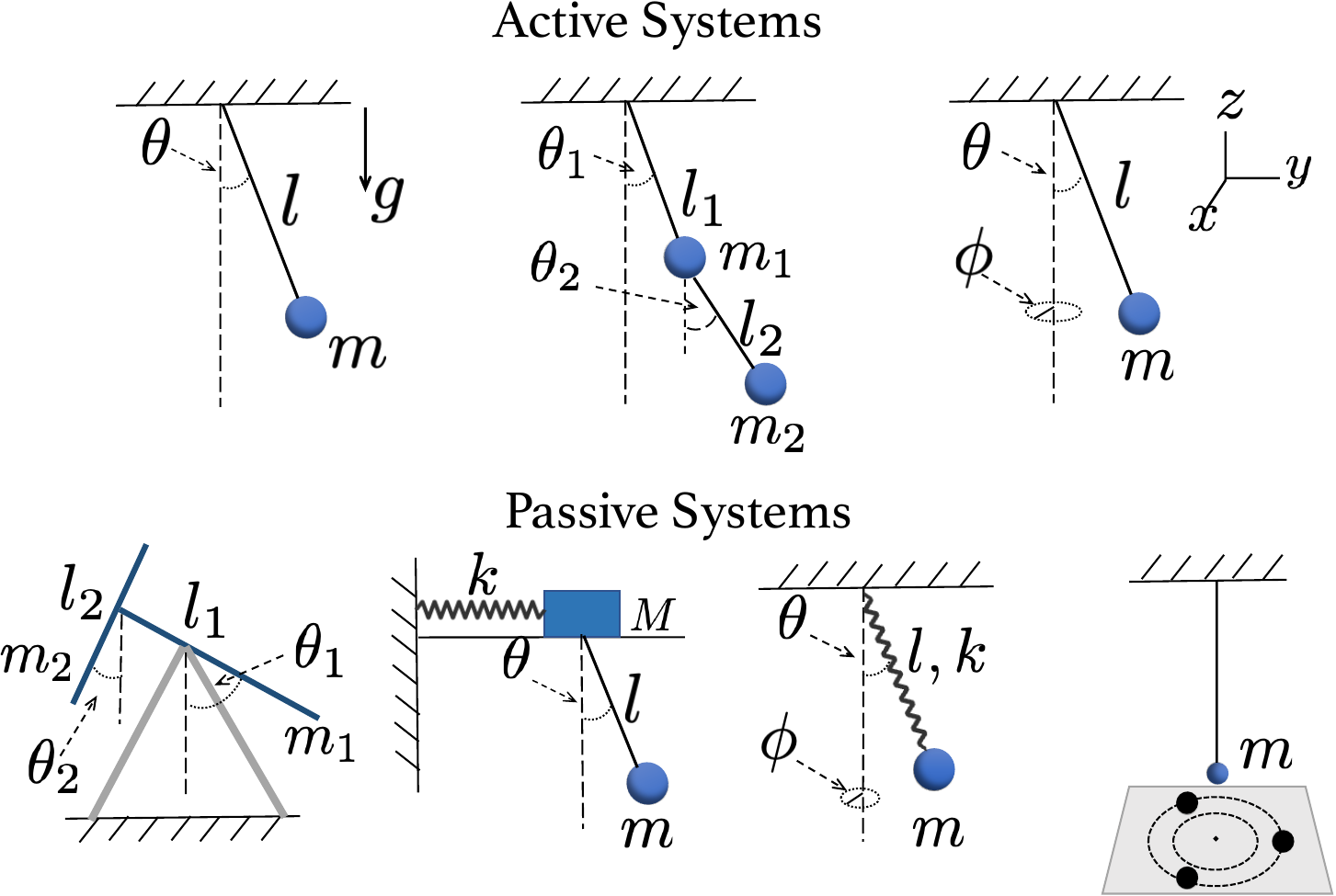}
  \caption{Two types of dynamical systems are considered, as shown from left to right: active systems, including  single pendulum,  double pendulum, and  spherical pendulum; and passive systems, including  chaotic pendulum,  cart-pendulum with a spring, spherical pendulum with a spring and magnetic pendulum.}
  \label{fig: twoTypeSystems}
\end{figure}

To evaluate the efficacy of our method, we conducted a series of comprehensive experiments on a wide range of ideal dynamical systems, as illustrated in Figure~\ref{fig: twoTypeSystems}.
The datasets capture essential second-order dynamical system behaviors, verifying the proposed modeling methodology.
These include both active and passive systems: the active systems consist of single pendulum, double pendulum, and spherical pendulum, while the passive systems include chaotic pendulum, cart-pendulum with a spring, spherical pendulum with a spring and magnetic pendulum. 
For Euler-Lagrange systems, active systems are characterized by the presence of external forces or control inputs $\boldsymbol{f}_{\text{ext}}$ . These external forces are typically applied to the generalized coordinates (degrees of freedom) of the system, enabling the modification of its dynamical behavior. In contrast, passive systems do not have external forces but leverage prior knowledge, e.g., gravitational potential energy, to model their dynamics.

\subsection{Experimental Setup}
This section details the experimental setup, including the datasets, baselines and metrics.

\textbf{Datasets.} 
We selected datasets consisting of second-order dynamical systems. All data were generated analytically by predefining ideal dynamical systems and simulating their trajectories using the fourth-order Runge-Kutta method.

\textbf{Baselines.}
To demonstrate the effectiveness of proposed method, we compared it with the following baselines:
{uDSR}~\cite{NEURIPS2022_dbca58f3}: A unified symbolic regression framework combines some strategies for complementary benefits.
{xL-SINDY}~\cite{purnomo2023sparse}: A work integrates SINDy with classical mechanics to identify sparse, interpretable expressions for Lagrangian dynamics.

\textbf{Evaluation metrics.} To evaluate the performance of our method, we adopt several metrics. Our objective is to accurately identify all relevant terms in the governing equations while minimizing the inclusion of irrelevant terms. The relative error, denoted as $\ell_2$, is defined as:
\(
\ell_2 = {\|\boldsymbol{\Lambda}_{\text{id}} - \boldsymbol{\Lambda}_{\text{true}}\|_2}/ {\|\boldsymbol{\Lambda}_{\text{true}}\|_2}
\), where $\boldsymbol{\Lambda}_{\text{id}}$ represents the identified coefficients and $\boldsymbol{\Lambda}_{\text{true}}$ represents the ground truth.
To address potential bias caused by significant disparities in coefficient magnitudes, we introduce a non-dimensional measure for a more balanced assessment.
The task of discovering governing equations can be viewed as a binary classification problem~\cite{rao2022discovering}, where the goal is to determine the presence or absence of each term in a candidate library. To this end, we employ precision ($P$) and recall ($R$) as evaluation metrics, defined as:
\(
P = \|\boldsymbol{\Lambda}_{\text{id}} \odot \boldsymbol{\Lambda}_{\text{true}}\|_0 / {\|\boldsymbol{\Lambda}_{\text{id}}\|_0}, \quad
R = {\|\boldsymbol{\Lambda}_{\text{id}} \odot \boldsymbol{\Lambda}_{\text{true}}\|_0} /{\|\boldsymbol{\Lambda}_{\text{true}}\|_0},
\)
where $\odot$ denotes element-wise product. A term is successfully identified if both $\boldsymbol{\Lambda}_{\text{id}}$ and $\boldsymbol{\Lambda}_{\text{true}}$ have non-zero entries.

These systems are analyzed under ideal conditions, where all rods except those in the chaotic pendulum are assumed to be massless. In the chaotic pendulum, the pivots of two rods are positioned differently along their robs, leading to variations in dynamical system's governing equation. 
The Euler-Lagrange equations for all dynamical systems discussed in this work are provided in Appendix~\ref{appendix:lagrangin function and energy}. 
The initial conditions for all experiments are randomly sampled from a uniform distribution within a specified range for each dynamical system, with each condition simulated for a duration of 20 seconds. The candidate library is constructed following the principle of minimal completeness, ensuring that redundant or interfering terms (e.g., simultaneously including $\sin^2\theta$ and $\cos^2\theta$) are excluded to maintain a concise and effective representation. 
We evaluated the proposed method using data corrupted by zero-mean white Gaussian noise $\mathcal{N}(0, \sigma)$ at varying noise levels. The noise was added exclusively to the measured data. For active systems, we assumed that the time-dependent external input is known. For passive systems, we incorporate prior knowledge of a constituent of the system by selecting one of the terms in the total Lagrangian as a known quantity. This allows us to constrain the optimization problem and improve the accuracy of the identified dynamics. 

\begin{table*} 
\centering
\footnotesize
\setlength{\tabcolsep}{1mm}
\begin{tabular}{cccccccccccccc}
\toprule
\multirow{2}{*}{Cases}&  \multicolumn{1}{c}{Metrics}  & \multicolumn{4}{c}{$\ell_2$ Error $\times 10^{-2}$} &\multicolumn{4}{c}{$P~(\%)$}&\multicolumn{4}{c}{$R~(\%)$} \\
\cmidrule(lr){3-6} \cmidrule(lr){7-10} \cmidrule(lr){11-14}
& Noise level &  0\% & 1\% & 10\% &avg. & 0\% & 1\% & 10\% &avg. & 0\% & 1\% & 10\% &avg.\\
\midrule
\multirow{3}{*}{Single pendulum}  
& uDSR  & {162.44} & {160.00} & 159.66   &   160.7     & {40} & {33.33} & {33.33} & 35.55    & {100} & {100} & {100}&   100\\
& xL-SINDy  & \textbf{0.31} & {1.53} & 26.72 &  9.52        & {100} & {100} & {100}&100      & {100} & {100} & {100}&  100\\
& Ours      & {0.41} & \textbf{1.31} & \textbf{10.94} &  \textbf{4.22}    & {100} & {100} & {100} & 100    & {100} & {100} & {100}&  100\\
\midrule
\multirow{3}{*}{Double Pendulum}   
& uDSR  & {119.40} & 106.42 & {105.48} &  110.43          & {37.50} & {27.27} & {22.22}& 29.00     & {60} & {40} & {20}&    40\\
& xL-SINDy  & \textbf{0.30} & 45.23 & {73.42}&  39.65           & {100} & {100} & {100} & 100    & {100} & {100} & {100}&100\\
& Ours      & {1.15} & \textbf{20.40} & \textbf{21.67} & \textbf{14.41}      & {100} & {100} & {100} & 100    & {100} & {100} & {100}    &100\\
\midrule
\multirow{3}{*}{Spherical pendulum}  
& uDSR  & {131.57} & 134.41 & {165.44}& 143.81         & {8.33} & {7.69} & {7.14} & 7.72    & {33.33} & {33.33} & {33.33} &33.33\\
& xL-SINDy  & \textbf{0.51} & 120.20 & {83.93} & 68.21        & {100} & {60} & {50} & 70    & {100} & {100} & {100} &100\\
& Ours      & {2.00} & \textbf{31.67} & \textbf{40.51} & \textbf{24.73}     & {100} & \textbf{100} & \textbf{100} & \textbf{100}    & {100} & {100} & {100} &100\\
\midrule
\multirow{3}{*}{Chaos pendulum} 
& uDSR  & 223.9 & 280.83 & 170.82 &  225.18              & {80} & {40} & {27.27} & 49.09    & {80} & {80} & {60}& 66.67\\
& xL-SINDy  & 0.51 & 5.62 & 9.12  &  5.08            & {100} & {83.33} & {87.50} & 90.28    & {100} & {100} & {85.71}& 95.24\\
& Ours      &  \textbf{0.16} & \textbf{2.10}  & \textbf{3.27}&  \textbf{1.84}     & {100} & \textbf{100} & \textbf{100} & \textbf{100}    & {100} & {100} & \textbf{100} &\textbf{100} \\
\midrule
\multirow{3}{*}{\begin{tabular}{@{}c@{}}Cart \\Spring Pendulum\end{tabular}}   
& uDSR  & {80.35} & {81.26} & {96.92}  & 86.18     & {75} & {75} & {27.27} & 59.09    & {75} & {75} & {42.86}&64.29\\
& xL-SINDy  & \textbf{1.34} & {1.55} & {16.10} &  6.33     & {100} & {87.50} & {87.50} & 91.67    & {100} & {100} & {100}&  100\\
& Ours      & {1.42} & \textbf{1.12}   & \textbf{2.42} &\textbf{1.65}      & {100} & \textbf{100} & {87.50}& \textbf{95.83}    & {100} & {100} & {100}&100\\
\midrule
\multirow{3}{*}{\begin{tabular}{@{}c@{}}Sphere \\Spring Pendulum\end{tabular}}  
& uDSR  & {135.28} & 241.79 & {333.51} & 236.86      & {22.22} & {1.92} & {2.50} & 8.88    & {57.14} & {14.29} & {14.29}& 28.57\\
& xL-SINDy  & \textbf{0.29} & 2.75 & {3.28} &  2.11     & {100} & {87.50} & {87.50}&  91.67    & {100} & {71.43} & {71.43}& 80.95\\
& Ours      & {0.39} & \textbf{2.60}  & \textbf{3.11}& \textbf{2.03}       & {100} & {87.50} & {87.50} &91.67     & {100} & \textbf{100} & \textbf{85.71}&\textbf{95.23}\\
\midrule
\multirow{3}{*}{Magnetical pendulum}  
& uDSR  & {314.15} & {322.40}  & 291.81   & 309.45      & {44.44}  &  40 & {30.77} & 38.40    & {57.14} & 57.14 & {57.14} &57.14\\
& xL-SINDy  & {1.56} & 14.33 & {34.22} &  16.70     & {100} & {100} & {75}&  91.67    & {100} & {85.71} & {85.71}&{90.47}\\
& Ours      & \textbf{1.37} & \textbf{3.60}  & \textbf{27.10} & \textbf{10.69}      & {100} & {100} & \textbf{100}& \textbf{100}     & {100} & \textbf{100} & \textbf{100}& \textbf{100}\\
\bottomrule
\end{tabular} 
\caption{
Performance comparison between our proposed framework and other methods in dynamical systems}
\label{Table: performance}
\end{table*}

\begin{table}[t!] 
\setlength{\tabcolsep}{1mm}
\centering
\footnotesize
\begin{tabular}{ccccccc}
\toprule
\multirow{1}{*}{Cases}&  \multirow{1}{*}{Groups} & \multicolumn{1}{c}{$\ell_2(\times 10^{-2})$} &\multirow{1}{*}{$P~(\%)$}&\multirow{1}{*}{$R~(\%)$} \\
\midrule
\multirow{3}{*}{Single Pendulum}  & A  & {0.41} & {100} & {100}\\
& B & 0.41 & {100} & {100} \\
& C & 12.36 & {33.33} & {100} \\
\midrule
\multirow{3}{*}{Double Pendulum}   & A & {1.15} & {100} & {100} \\
& B  & 1.15 & 100 & {100} \\
& C  & 58.65 & 83.33 & {60} \\
\midrule
\multirow{3}{*}{Spherical Pendulum}  & A  & {2.00} & {100} & {100} \\
& B  & 2.04 & 100 & {100} \\
& C  & 91.55 & 50 & {100} \\
\midrule
\multirow{3}{*}{Chaos Pendulum} & A  & {0.16} & {100}  & {100}\\
& B  & 0.16 & 100 & 100 \\
& C  & 18.25 & 87.50 & 85.71 \\
\midrule
\multirow{3}{*}{\begin{tabular}{@{}c@{}}Cart \\Spring Pendulum\end{tabular}}   & A & {1.42} & {100}   & {100}\\
& B & 1.40 & 100 & {100} \\
& C & 47.25 & 87.50 & {75} \\
\midrule
\multirow{3}{*}{\begin{tabular}{@{}c@{}}Spherical \\Spring Pendulum\end{tabular}} & A & 0.39 & 100 & 100 \\
& B & 1.13 & 100 & 100 \\
& C & 8.51 & 85.70 & 85.70 \\
\midrule
\multirow{3}{*}{Magnetical Pendulum}  & A  & {1.37} & {100}  & {100} \\
& B  & 3.01 & 100 & {100} \\
& C  & 40.55 & 75 & {85.71} \\
\bottomrule
\end{tabular} 
\caption{Performance under varying experimental scenarios}
\label{Table: ablationStudy}

\end{table}


\subsection{Main Results}
\begin{figure}[htbp]
    \centering
        \includegraphics[width=0.98\linewidth]{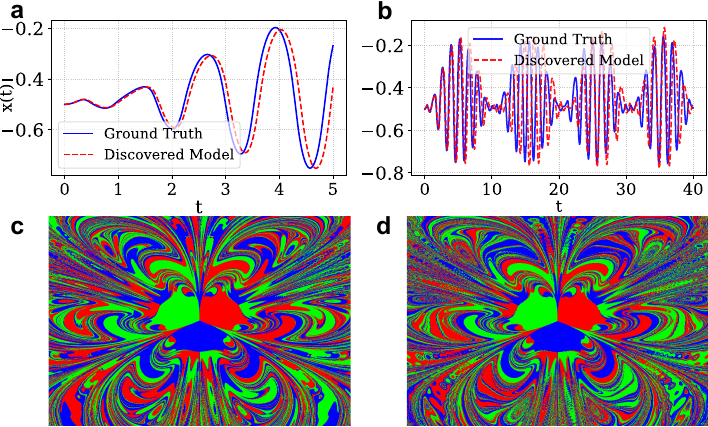}
        \caption{
        \textbf{a} Matches short-term;
\textbf{b} Diverges long-term but preserves key statistical properties(amplitude range, oscillations).
        The basins of attraction for three magnets (colored red, blue, and green) are distributed along the edges of an equilateral triangle. \textbf{c} shows the true basin of attraction for magnetic pendulum, while \textbf{d} displays the basin obtained with the discovered parameters of system.
        }
        \label{fig:basin_compare}
\end{figure}

\paragraph{Discovery results.} Based on our evaluation metrics, a detailed analysis of the experimental results achieved by our method is provided in Table~\ref{Table: performance}. We evaluate our method in comparison with xL-SINDy and uDSR as baseline models. The results listed are averaged over five independent trials. The evaluation metrics computed from the discovered Lagrangians of different dynamical systems under various noise levels are provided. The better performing values under the same metrics are highlighted in bold. Experimental results show that the performance of both models is nearly identical under zero-noise measurement conditions. Although the ($\ell_2$) still exhibit small errors, we consider these errors to be within a reasonable range. Under higher noise levels, our model demonstrates significantly lower relative error ($\ell_2$) and achieves competitive performance in terms of precision ($P$) and recall ($R$).
Overall, our method exhibits better robustness and accuracy under noise conditions.
Using the magnetical pendulum as an example, we validate our dynamical model by comparing its predictions with the real system (Figure~\ref{fig:basin_compare}\textbf{ab}). 
Due to chaotic divergence, precise long-term prediction is fundamentally impossible. Although trajectories eventually diverge, our model preserves key statistical properties (amplitude and frequency), maintaining behavior similar to the real system.
After equation discovery, we incorporate a damping term to visualize the basin of attraction diagram. It shows a close match in capturing the chaotic dynamics.
Figure~\ref{fig:basin_compare}\textbf{cd} compares the true basin of attraction with the basin from our discovered model, showing accurately captured chaotic dynamics. 
Furthermore, our approach demonstrates exceptional suitability for scenarios involving randomly sampled or incomplete data, a common challenge in real-world applications. To rigorously evaluate the robustness of our method against data imperfections, we conducted a comprehensive set of comparative experiments. The experimental design consisted of two distinct groups: Group A utilized the complete time-series dataset, while Group B employed a dataset with 5\% randomly missing samples, where the missing points were uniformly distributed across the temporal dimension. The experimental results, as detailed in Table~\ref{Table: ablationStudy}, reveal that the presence of missing data has a statistically negligible impact on the performance metrics of our method, with performance degradation within an acceptable range across evaluated metrics. acceptable range across evaluated metrics.acceptable range across evaluated metrics.
This reflects the robustness of B-spline curves against noise and their flexibility in handling time-series data without requiring strict dependencies. Specifically, B-splines' local support property allows them to adapt to irregularities in the data, such as missing points, without significantly affecting the overall fit. Additionally, their smoothness and differentiability ensure that the reconstructed dynamics remain physically plausible, even under incomplete or noisy conditions. These properties make B-splines particularly suitable for applications where data quality may vary or where strict temporal alignment is challenging to achieve.

\begin{figure}[t]
  \centering
  \includegraphics[width=0.98\linewidth]{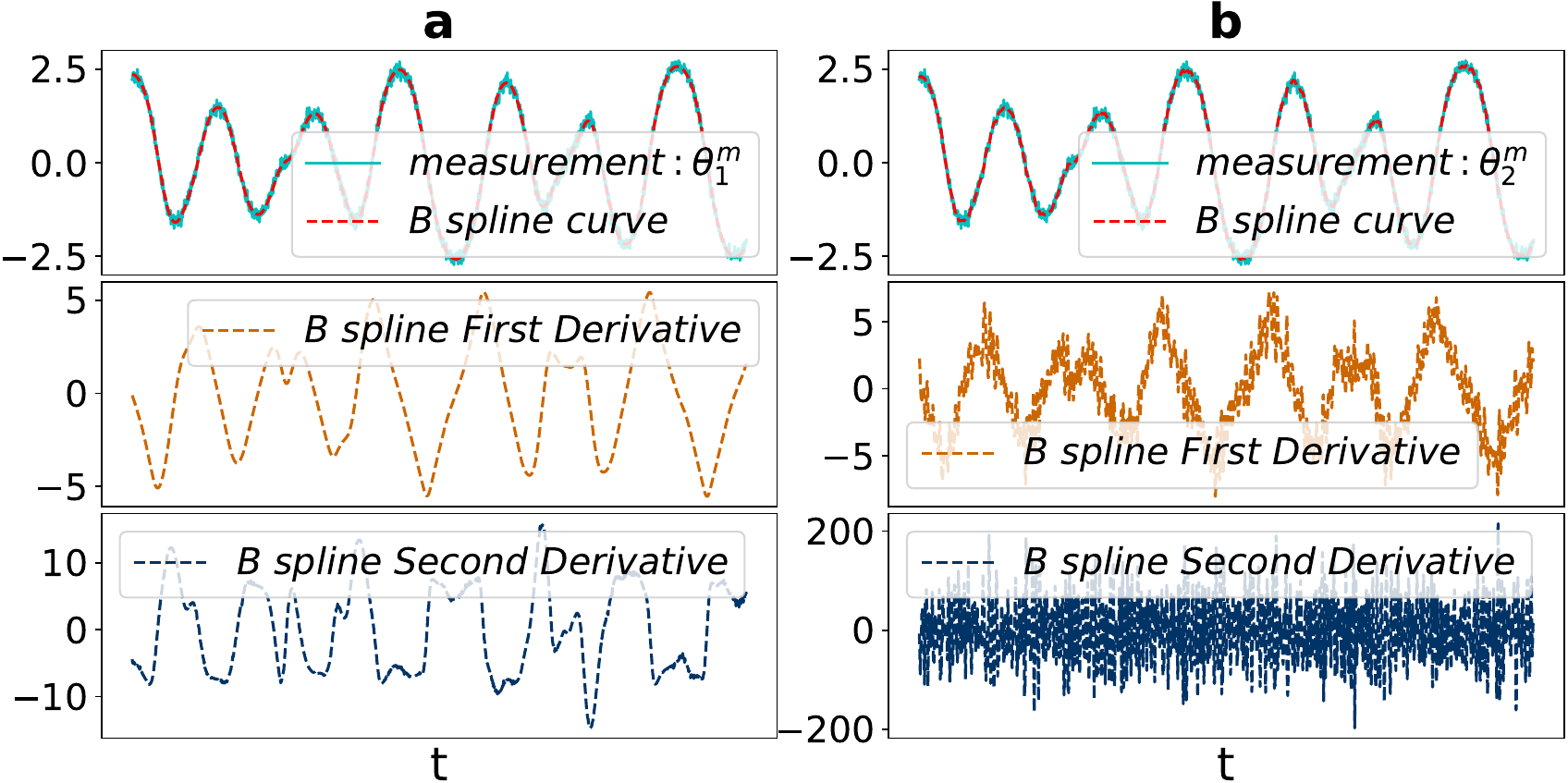}
  \caption{
  Impact of Second-Order Derivative Regularization on Performance.
\textbf{a}: Performance of our proposed model in fitting the measurements and their derivatives. \textbf{b}: Performance of the model without the regularization term.
  }
  \label{fig: Ablation}
\end{figure}

\paragraph{Ablation Study.}
We conducted an ablation study to validate the effectiveness of the module that imposes constraints on higher-order derivatives of the system's degrees of freedom. This module is designed to mitigate noise amplification caused by chain rule differentiation, which is particularly critical in second-order dynamical systems. To evaluate its impact, we introduced an ablation model that excludes the regularization term on the system's second-order derivatives. Using the chaotic pendulum as a representative example, we present the training results of the ablation model in Figure~\ref{fig: Ablation}, where \textbf{a} represents our proposed method and \textbf{b} represents the model without the regularization term on second-order derivatives.
Our observations reveal that, in the absence of the regularization term, the data loss of the system is minimized (the B-spline curve achieves near-perfect fitting of the measurement points). However, the first and second derivatives of the time-dependent functions for the degrees of freedom are poorly approximated, indicating a failure to capture the underlying dynamics accurately. Specifically, the second derivatives exhibit significant deviations from the ground truth, leading to a substantial deterioration in both training speed and model performance. This highlights the importance of constraining higher-order derivatives to ensure physical consistency and robustness. 
As shown in Table~\ref{Table: ablationStudy}, Group A represents the full model, while Group C (without regularization) exhibits degraded performance, validating the necessity of regularization.
These results demonstrate that our proposed framework, which incorporates regularization on higher-order derivatives, is essential for accurate equation discovery, especially under conditions of incomplete or noisy data. By enforcing reasonable constraints on the system's dynamics, our approach effectively balances data fitting and physical plausibility, enabling reliable identification of governing equations.

\paragraph{Discussion and Limitation.}

Through our proposed differentiable sparse identification of Lagrangian dynamics, we leverage the differentiability of B-spline basis functions to fit curves and their derivatives by optimizing control points. Using only intuitive measurement data, our method discovers Lagrangian dynamics equations for various systems, effectively mitigating noise amplification caused by chain rule differentiation on candidate function libraries. This enables a more accurate identification of target equations.  

Although our method performs well in identifying Lagrangian dynamics equations, it still has some limitations. When dealing with complex systems involving multiple degrees of freedom and high coupling, the size of the candidate function library increases significantly, leading to a sharp rise in computational complexity. This not only prolongs training time but may also make it difficult to converge to the correct solution due to combinatorial explosion. Moreover, although B-splines exhibit strong noise resistance, the computation of higher-order derivatives may still be affected under extreme noise conditions, impacting the model's stability. Future work could explore more efficient library construction methods, incorporating domain knowledge or adaptive strategies, to further enhance the model's applicability and robustness in complex systems.

\section{Conclusion}

This work introduces a novel differentiable framework for the sparse identification of Lagrangian dynamics from observational data. By integrating cubic B-spline approximation with sparse regression, our method effectively addresses key limitations in existing approaches, particularly in modeling complex nonlinearities and mitigating the impact of noise.
The core of our framework is a novel integration of cubic B-splines into the system identification process, enabling flexible and accurate modeling of complex dynamics. This approach incorporates a robust equation discovery mechanism that efficiently utilizes limited measurements and leverages physical constraints. Furthermore, we introduce a recursive derivative computation scheme designed specifically to reduce noise sensitivity in second-order dynamical systems.
Experimental results demonstrate that our method significantly outperforms baseline approaches in both accuracy and robustness, particularly in scenarios with high noise levels or sparse data. By leveraging the recursive properties of B-spline basis functions, our framework achieves a precise reconstruction of the governing equations while preserving their physical plausibility. 
This advancement enhances the reliability of data-driven discovery in complex mechanical systems and paves the way for hybrid models that fuse empirical insights with established physical principles. This synthesis yields actionable understanding of nonlinear dynamics across scientific and engineering fields.

\section{Acknowledgments}
The work is supported by the National Natural Science Foundation of China (No. 92270118 and No. 62276269) and the Beijing Natural Science Foundation (No. 1232009). 

\bibliography{aaai2026}

\newpage
\appendix

\onecolumn

\setcounter{figure}{0}
\setcounter{table}{0}
\setcounter{equation}{0}
\renewcommand{\thefigure}{S\arabic{figure}}
\renewcommand{\thetable}{S\arabic{table}}
\renewcommand{\theequation}{S\arabic{equation}}

\noindent\textbf{\large APPENDIX}

The appendix contains supplementary derivations and implementation details to support our method.

\setcounter{secnumdepth}{1}
\section{Derivation of the Second Derivative of Cubic B-spline Basis Functions}
\label{app:sec_derivative}

The basis functions $N_{i, p}(u)$ are defined as:
\begin{equation}
N_{i, 0}(u)= \begin{cases}1 & \text { if } u_i \leq u<u_{i+1} \\ 0 & \text { otherwise }\end{cases}.
\end{equation}
and for degree $p \geq 1$ :
\begin{equation}
N_{i, p}(u)=\frac{u-u_i}{u_{i+p}-u_i} N_{i, p-1}(u)+\frac{u_{i+p+1}-u}{u_{i+p+1}-u_{i+1}} N_{i+1, p-1}(u).
\end{equation}
The first derivative $N_{i, p}^{'}(u)$ is obtained by differentiating the recursion:

\begin{equation}
N_{i, p}^{'}(u) = \frac{d N_{i, p}(u)}{d u}=\frac{p}{u_{i+p}-u_i} N_{i, p-1}(u)-\frac{p}{u_{i+p+1}-u_{i+1}} N_{i+1, p-1}(u).
\end{equation}
The second derivative $N_{i, p}^{''}(u)$ is derived by differentiating the first derivative:
\begin{equation}
        \begin{aligned}
            \frac{d^2 N_{i, p}(u)}{d u^2}= &  \frac{d}{d u}\left(\frac{p}{u_{i+p}-u_i} N_{i, p-1}(u)\right)-\frac{d}{d u}\left(\frac{p}{u_{i+p+1}-u_{i+1}} N_{i+1, p-1}(u)\right) \\
            = & \frac{p}{u_{i+p}-u_i} \cdot \frac{d N_{i, p-1}(u)}{d u}  -\frac{p}{u_{i+p+1}-u_{i+1}} \cdot \frac{d N_{i+1, p-1}(u)}{d u} \\
            = & \frac{p}{\left(u_{i+p}-u_i\right)^2} N_{i, p-1}(u)+\frac{p}{u_{i+p}-u_i} \frac{d N_{i, p-1}(u)}{d u} \\
            &-\frac{p}{\left(u_{i+p+1}-u_{i+1}\right)^2} N_{i+1, p-1}(u)-\frac{p}{u_{i+p+1}-u_{i+1}} \frac{d N_{i+1, p-1}(u)}{d u}
        \end{aligned}
\end{equation}
Then, we continue to expand the equation.
\begin{equation}
\begin{aligned}
    \frac{d N_{i, p-1}(u)}{d u} & =\frac{p-1}{u_{i+p-1}-u_i} N_{i, p-2}(u)-\frac{p-1}{u_{i+p}-u_{i+1}} N_{i+1, p-2}(u) \\
    \frac{d N_{i+1, p-1}(u)}{d u} & =\frac{p-1}{u_{i+p}-u_{i+1}} N_{i+1, p-2}(u)-\frac{p-1}{u_{i+p+1}-u_{i+2}} N_{i+2, p-2}(u)
\end{aligned}
\end{equation}
Finally, through simplification, we arrive at the final expression:
\begin{equation}
    \begin{aligned}
         \frac{d^2 N_{i, p}(u)}{d u^2} = &\frac{p(p-1)}{\left(u_{i+p}-u_i\right)\left(u_{i+p-1}-u_i\right)} N_{i, p-2}(u)\\
         & - \frac{p(p-1)}{\left(u_{i+p}-u_i\right)\left(u_{i+p}-u_{i+1}\right)} N_{i+1, p-2}(u) \\
        & -\frac{p(p-1)}{\left(u_{i+p+1}-u_{i+1}\right)\left(u_{i+p}-u_{i+1}\right)} N_{i+1, p-2}(u) \\
        & +\frac{p(p-1)}{\left(u_{i+p+1}-u_{i+1}\right)\left(u_{i+p+1}-u_{i+2}\right)} N_{i+2, p-2}(u)
    \end{aligned}
\end{equation}%
Substituting $p=3$ into the expression yields the desired result.

\section {The Lagrangian equations for all dynamical systems}
\label{appendix:lagrangin function and energy}

To ensure consistent experimental conditions, all simulations in this study are conducted on an Intel Core i9-13900 CPU workstation with a NVIDIA GeForce RTX 4090 GPU. Initial conditions were randomly generated for all equations.

\subsection{Single pendulum}
The Euler-Lagrange equation for the pendulum system is given below:
\begin{equation}
    \mathcal{L} = \frac{1}{2} m l^2 \dot{\theta}^2 + m g l \cos\theta.
\end{equation}

The Lagrangian $\mathcal{L}$ for the single pendulum system comprises two terms: the kinetic energy term $\frac{1}{2}ml^2\dot{\theta}^2$ where $m$ is the mass, $l$ is the pendulum length, and $\dot{\theta}$ is the angular velocity; and the potential energy term $-mgl\cos\theta$ where $g$ is gravitational acceleration and $\theta$ is the angle. 

\subsection{Double pendulum}
The Euler-Lagrange equation for double pendulum system can be described by the equation:
\begin{equation}
    \begin{aligned}
        \mathcal{L}=& \frac{1}{2} m_1 l_1^2 \dot{\theta}_1^2+\frac{1}{2} m_2\left(l_2^2 \dot{\theta}_2^2+2 l_1 l_2 \dot{\theta}_1 \dot{\theta}_2 \cos \left(\theta_1-\theta_2\right)+l_1^2 \dot{\theta}_1^2\right)\\
        &+\left(m_1+m_2\right) g l_1 \cos \theta_1+m_2 g l_2 \cos \theta_2,
    \end{aligned}
\end{equation}
where $m_1$ and $m_2$ represent the masses of the two connected pendulums with lengths $l_1$ and $l_2$ respectively, $\dot{\theta}_1$ and $\dot{\theta}_2$ are their angular velocities, and the coupling term $2l_1l_2\dot{\theta}_1\dot{\theta}_2\cos(\theta_1-\theta_2)$ captures their interaction through the angle difference $(\theta_1-\theta_2)$. The potential energy terms now include contributions from both pendulums.

\subsection{Spherical pendulum} 
The Euler-Lagrange equation for the Spherical pendulum is:
\begin{equation}
    \mathcal{L}=\frac{1}{2} m l^2\left(\dot{\theta}^2+\sin ^2 \theta \dot{\phi}^2\right)+m g l \cos \theta,
\end{equation}
where $m$ is mass, $l$ is pendulum length,  $\dot{\theta}$ is polar angular velocity, $\dot{\phi}$ is azimuthal angular velocity, and $\theta$ is the polar angle.

\subsection{Chaos pendulum} 
In the chaotic pendulum system, the pivot point of the first rod is located at $1/2$ of its length, while the pivot point of the second rod is located at $1/3$ of its length. This asymmetric pivot configuration leads to more complex dynamical behavior and significantly influences the formulation of the dynamical equations.
The chaotic pendulum's Lagrangian is
\begin{equation}
    \begin{aligned}
        \mathcal{L}=& \frac{m_1 l_1^2 \dot{\theta}_1^2}{24} + \frac{m_2 l_1^2 \dot{\theta}_1{ }^2}{8}+\frac{m_2 l_2^2 \dot{\theta}_2^2}{18}+\frac{m_2 l_1 l_2 \dot{\theta}_1 \dot{\theta}_2  \cos \left(\theta_1-\theta_2\right)}{12} \\
        & + \frac{1}{2} m_2 g l_1  \cos \left(\theta_1\right) + \frac{1}{6} m_2 g  l_2 \cos \left(\theta_2\right),
    \end{aligned}
\end{equation}
where $m_1$ and $m_2$ represent the masses of the two connected pendulums with lengths $l_1$ and $l_2$ respectively, ${\theta}_1$ and ${\theta}_2$ are their angle.

\subsection{Cart spring pendulum}
The system's Lagrangian is shown as follow:
\begin{equation}
    \mathcal{L}= \frac{1}{2} M \dot{x}^2+\frac{1}{2} m\left(\dot{x}^2+2 \dot{x} l \dot{\theta} \cos (\theta)+l^2 \dot{\theta}^2\right)-m g l(1-\cos (\theta))- \frac{1}{2}k(x-d)^2, 
\end{equation}
where $M$ is the mass of the cart, $x$ is the position of the cart, $\theta$ is the angle of the spring, $k$ is the stiffness of the spring and $d$ is its equilibrium length.

\subsection{Spherical spring pendulum}
The spherical spring pendulum's Lagrangian $\mathcal{L}$ contains three key components: (1) kinetic energy with $\dot{x}$ (radial velocity), $\dot{\theta}$ (polar angular velocity), and $\dot{\phi}$ (azimuthal angular velocity); (2) gravitational potential depending on $x$ (spring length) and $\theta$ (polar angle); (3) spring potential with $k$ (spring constant) and $d$ (equilibrium length). Mass $m$ scales all terms.
\begin{equation}
    \mathcal{L}=  
    \frac{1}{2} m\left(\dot{x}^2+x^2 \dot{\theta}^2+x^2 \sin ^2 \theta \dot{\phi}^2\right)+m g x \cos \theta-\frac{1}{2} k(x-d)^2.
\end{equation}

\subsection{Magnetical pendulum}

The Euler-Lagrange equation for Magnetical pendulum system can be expressed as:

\begin{equation}
    \mathcal{L}=  
    \frac{1}{2} m\left(\dot{x}^2+\dot{y}^2\right)+\sum_{i=1}^N \frac{C_i}{\sqrt{\left(x_i-x\right)^2+\left(y_i-y\right)^2+d^2}}-\frac{1}{2} C\left(x^2+y^2\right),
\end{equation}
where 
\( C \) is proportional to the effects of gravity, \(C_i\) is a constant depending on the magnetic moments and the permeability of free space, \( N \) is the number of magnets, the \( i \)-th magnet is positioned at \((x_i, y_i)\), \( d \) is the distance between the pendulum at rest and the plane of magnets. Additionally, we assume that the pendulum length is long compared to the spacing of the magnets. Thus, we may assume for simplicity that the metal ball moves about on an \( xy \)-plane.
For simplicity, we neglect the damping term, and the system is conservative, meaning energy is conserved. 


\end{document}